\pdfoutput=1 
\documentclass{llncs}
\usepackage{booktabs}
\usepackage{xcolor}
\usepackage{amsmath,amssymb,amsfonts}
\usepackage{graphicx}
\usepackage{verbatim}
\usepackage[ruled,vlined, linesnumbered]{algorithm2e}
\usepackage{url}

\DeclareMathOperator*{\argmin}{arg\,min}
\usepackage{dsfont}
\newcommand{\bftab}{\fontseries{b}\selectfont}
\usepackage{bold-extra}
\usepackage{todonotes}
\usepackage{float}
\usepackage{hyperref}

\begin{document}
\title{R-GCN: The R Could Stand for Random}
\author{Vic Degraeve \and 
 Gilles Vandewiele \and
 Femke Ongenae \and
 Sofie Van Hoecke}
\authorrunning{Degraeve, Vandewiele, Ongenae and Van Hoecke}
%
 \institute{IDLab, Ghent University - imec \\
 \email{\{firstname\}.\{lastname\}@ugent.be}}

\maketitle             

\begin{abstract}
The inception of the Relational Graph Convolutional Network (\textsc{r-gcn}) marked a milestone in the Semantic Web domain as a widely cited method that generalises end-to-end hierarchical representation learning to Knowledge Graphs (KGs). \textsc{r-gcn}s generate representations for nodes of interest by repeatedly aggregating parameterised, relation-specific transformations of their neighbours. However, in this paper, we argue that the the \textsc{r-gcn}'s main contribution lies in this ``message passing'' \textit{paradigm}, rather than the learned weights. To this end, we introduce the ``Random Relational Graph Convolutional Network" (\textsc{rr-gcn}), which leaves \textit{all} parameters untrained and thus constructs node embeddings by aggregating \emph{randomly} transformed \textit{random} representations from neighbours, i.e., with no learned parameters. We empirically show that \textsc{rr-gcn}s can compete with fully trained \textsc{r-gcn}s in both node classification and link prediction settings.
\end{abstract}

\section{Introduction}
Knowledge Graphs (KGs) are the ideal data structure to represent both expert knowledge and observational data, and can be used to reveal new insights about the modelled domain. The latter becomes even more true thanks to today's hybrid Machine Learning (ML) methodologies~\cite{ristoski2016semantic,palmonari2020knowledge} where semantically enriched data is combined with a statistical method's ability to learn. As these models typically operate on Euclidean data, the integration of KGs into their decision making processes involves a non-trivial transformation from information represented as a variable number of nodes and edges to fixed size numerical vectors, i.e.,~graph embeddings. Relational Graph Convolutional Networks (\textsc{r\mbox{-}gcn}s) embed semantic information about an entity contained in a KG by iteratively updating node representations. First, the previous representations for a node's neighbours are collected and passed through a learned, relation-specific, transformation. These transformed representations are referred to as ``messages''. Second, an aggregation of the collected messages and the node's own state yields the updated node representation~\cite{battaglia2018relational}.

In this work, we ablate the parameterisation of the \textsc{r-gcn}'s message passing step by randomly initialising and \textit{freezing} the relation-specific transformations.
We evaluate these ``Random Relational Graph Convolutional Networks'' (\textsc{rr\mbox{-}gcn}s) on nine node classification datasets, and empirically show that these random transformations produce embeddings for node classification that are on par with -- and sometimes even outperform -- those produced by \textsc{r-gcn}s. We also compare the embeddings produced by our \textsc{rr-gcn}s to an end-to-end trained \textsc{r-gcn} for a link prediction task and obtain very similar results. 

The remainder of the paper is structured as follows. In Section~\ref{sec:contributions}, we summarise our main contributions to the field of KG embedding. Section~\ref{sec:background} provides the necessary background in concepts that are fundamental to our \textsc{rr-gcn}. Next, in Section~\ref{sec:method}, we discuss our technique formally. We then compare \textsc{rr-gcn} to the original \textsc{r-gcn} and discuss the implications of these results in Section~\ref{sec:results}. Finally, we provide some closing remarks in Section~\ref{sec:conclusion} and discuss future work in~\ref{sec:futurework}.

\section{Contributions}
\label{sec:contributions}
This paper is the first to evaluate \emph{fully random} \textsc{r-gcn}s for KG embeddings. We show that random transformations can capture a surprising amount of information. On the one hand, this makes our \textsc{rr-gcn}s an interesting baseline model when developing trained embedding methods. On the other hand, the effectiveness of these random transformations illustrates that, for KGs, the \textsc{r-gcn}'s message passing and aggregation \textit{paradigm} is more significant than the actual parameters, which have to be obtained through an expensive training procedure. This opens up avenues for further research on more efficient, and more powerful, message passing parameretisations for KGs. 

\section{Background} \label{sec:background}
In this section, we touch upon the concepts that are fundamental to our \textsc{rr-gcn} approach: (i) we briefly outline prior research directions in Knowledge Graph Embedding (KGE); (ii) we provide the necessary theory of \textsc{r-gcn}s; and (iii) we discuss the concept of learning from random transformations.

\subsection{Representation Learning for KGs}
Techniques to embed substructures in KGs can be categorised into four groups. A first category compromises of techniques that extract generic properties from neighbourhoods of substructures of interest. These can either be feature-based~\cite{Steenwinckel2021INK,vandewiele2020mindwalc} or exploit similarities with other substructures of interest, i.e., kernel functions~\cite{de2013fast,vishwanathan2010graph}. A second category consists of (algebraic) embedding spaces learned using tensor factorisation or through negative sampling~\cite{bordes2013translating,kazemi2018simple,yang2015embedding,dettmers2018convolutional,zhang2019quaternion}. A third category adapts existing natural language processing (NLP) techniques, such as Word2Vec~\cite{mikolov2013efficient}, to graph structures~\cite{ristoski2016rdf2vec}. A fourth and final category contains the message passing architectures that are trained end-to-end to aggregate relevant information around the substructures of interest~\cite{schlichtkrull2018modeling,2019transgcn,nathani2019learning,compgcn}. \textsc{r-gcn}s belong to this final category.

\subsection{R-GCN}
Graph Neural Networks (\textsc{gnn}s) allow for graph ML by learning how to update a node's representation based on its neighbours. These \textsc{gnn}s work analogously to classical neural networks: they start from initial node features which are passed through multiple \textsc{gnn} layers to refine and abstract the representations by mixing in information from one additional hop with every layer.

A single \textsc{gnn} layer operates in three steps: every node (1) generates a ``message'' based on its current representation and sends it along its outgoing edges, (2) aggregates incoming messages, and (3) updates its representation based on the aggregated messages and its own previous representation.

The definitions of the message, aggregation and update functions differentiate several subtypes of \textsc{gnn}s. The most prominent of these subtypes, the Graph Convolutional Network (\textsc{gcn})~\cite{kipf2016semi}, uses a learned transformation matrix as its update function, aggregates by averaging messages (including the node's own message) and passes the aggregations through an activation function to yield new node representations (see Equation \ref{eq:methgcn}).
\begin{equation}
    \label{eq:methgcn}
    h_i^{(l+1)} = \phi \Bigg(\sum_{j \in \mathcal{N}(i) \mathsmaller{\cup}  i} \frac{1}{|\mathcal{N}(i)| + 1} \;\, W^{(l)} h_j^{(l)} \Bigg)
\end{equation}
Here, $\phi$ denotes the activation function used, $h_i^{(l)}$ is the representation for node $i$ at layer $l$, $\mathcal{N}(i)$ is the set of neighbours for node $i$, and $W^{(l)}$ is the learned transformation matrix in layer $l$. This formulation assumes unweighted edges.

\textsc{r-gcn}s extend \textsc{gcn}s to support typed edges. As KGs are multi-relational, this extension is particularly interesting for hybrid ML. They are almost identical to standard \textsc{gcn}s but learn a different transformation matrix $W_r^{(l)}$ for every relation $r \in \mathcal{R}$. A separate transformation $W_0^{(l)}$ takes care of self-loops (see Equation~\ref{eq:methrgcn}). Note that $\mathcal{R}$ contains two copies of every relation in the original KG, as \textsc{r-gcn}s also take outgoing (i.e. inverse) edges into account, with different learned transformations.
\begin{equation}
    \label{eq:methrgcn}
    h_i^{(l+1)} = \phi \Bigg(W_0^{(l)} h_i^{(l)} + \sum\limits_{r \in \mathcal{R}}\sum_{j \in \mathcal{N}_r(i)} \frac{1}{|\mathcal{N}_r(i)|} \;\, W_r^{(l)} h_j^{(l)} \Bigg)
\end{equation}
Learning separate transformations per relation results in extra parameters and might not work well for sparse types. Weight sharing between relations has been proposed to reduce overfitting and the number of required parameters~\cite{schlichtkrull2018modeling}. Using a basis decomposition, a layer's parameters are reduced to a fixed set of base parameter matrices $V_i$ (Equation~\ref{eq:basisdecomp}).
\begin{equation}
    \label{eq:basisdecomp}
    W^{(l)}_{r} = \sum_{b=1}^{B}a^{(l)}_{rb}V^{(l)}_{b}
\end{equation}

\subsection{Learning from Random Representations}
Our approach is inspired by an algorithm from the time series classification domain, \textsc{rocket}~\cite{dempster2020rocket,dempster2021minirocket}. This algorithm applies a large number of random convolutions and aggregations to the raw time series input to extract features with state-of-the-art predictive performance. These random transformations do not have to be learned and can be applied very quickly, making \textsc{rocket} more scalable than other time series classification techniques. Two types of aggregations are used within \textsc{rocket}: (i) max-pooling, which just retrieves the maximum value of a certain convolutional filter when slided across the time series, and (ii) the proportion of positive values (\textsc{ppv}), which, for every filter, captures the proportion of the input time series for which the output of the convolution operation is positive. We will discuss the latter aggregation method in more depth in the context of graphs in Section~\ref{subsec:ppv}.

The idea of freezing, or not training, layers in neural networks is not novel either. Reservoir Computing~\cite{reservoir} is a paradigm in recurrent neural networks where inputs are first passed through any black-box non-linear system called a ``reservoir", which could be an untrained neural network. As such, only the final layer that maps the learned representations to the target output is trained.
The paradigm of random modelling often introduces a trade-off between efficiency and effectiveness. As many random, albeit less effective, transformations can often be applied very efficiently, they allow to scale to higher numbers. An example of a technique where quantity matters over quality is ExtraTrees~\cite{geurts2006extremely}, where a large number of decision trees with random splits yield good results.

In a blog post, detailing the inner workings of (non-relational) \textsc{gcn}s~\cite{kipfblog}, Thomas Kipf hinted at the discriminative power of representations resulting from untrained transformations, but this hypothesis was never formally evaluated, let alone for KGs. In a recent paper~\cite{chian2021learning}, random transformations were evaluated in the context of link prediction in a neuromorphic computing setting. In this research, the authors backpropagated through random weights to train initial node embeddings. Other research demonstrated that keeping the initial node features random and frozen, and only training the message passing transformations results in good performance for unirelational graphs~\cite{surprising,randomfeatures}. For KGs, it is even possible to deterministically generate meaningful node features, avoiding the need to learn initial embeddings in the first place~\cite{nodepiece}.
As will be explained in the remainder of this paper, we are the first to explore \emph{fully} random networks for KGs, that keep both the initial representations and the transformations frozen.

\section{Methodology} \label{sec:method}
In this section, we elaborate on the modifications we made to the original \textsc{r-gcn} algorithm. We also present a new information aggregation function for graphs: \textsc{ppv}. Finally, we compare memory requirements for trained and random \textsc{r-gcn}s.

\subsection{Modifying the R-GCN Layer}
Our \textsc{rr-gcn} layers use the same message paradigm as \textsc{r-gcn}s (see Equation~\ref{eq:methrgcn}). The most important difference is that we randomly initialise all transformation matrices using the Glorot uniform initialisation function~\cite{glorot} and then keep them fixed to freeze the network. Basis decomposition is no longer required and thus not used, as overfitting cannot occur since no training is involved.

In~\cite{schlichtkrull2018modeling}, the node features that are fed to the first \textsc{r-gcn} layer are one-hot encoded. Using one-hot node features effectively assigns a separate initial embedding matrix $W^{(0)}_r$ per relation type in the first message passing layer. This is indeed desirable for \textsc{r-gcn}s, since the output node representation dimensionality is usually very small ($10$ to $16$ in the original paper~\cite{schlichtkrull2018modeling}). Starting with small, randomly initialised, node representations limits the level of detail in the initial node characterisations and leads to worse performance for some datasets~\cite{thanapalasingam2021relational}. 
As our \textsc{rr-gcn}s use much larger node representations, we can feed our first message passing layer random features $h^{(0)}_{i} \in \mathbb{R}^{d}$, with $d$ the embedding size, -- which we also do not train -- and save them as a single seed. In our experiments, this did not impact performance compared to one-hot encoded inputs.

\subsection{Proportion of Positive Values (PPV)} \label{subsec:ppv}
The performance of \textsc{rocket}'s random feature extraction for time series depends strongly on their use of \textsc{ppv}~\cite{dempster2020rocket} pooling, as is also apparent from their second version which abandons max-pooling and uses \textit{only} \textsc{ppv} features~\cite{dempster2021minirocket}. Inspired by their success, we adapted a variant of \textsc{ppv} to graphs that aims to encapsulate additional information about a node's neighbourhood. Given a matrix of node representations $H$, we define \textsc{ppv} as the proportion of strictly positive values in a 1-hop neighborhood per representation dimension (see Equation \ref{eq:ppv}).
\begin{equation}
    \label{eq:ppv}
    P = \textsc{ppv}(H, \mathcal{N}) = \sum\limits_{j \in \mathcal{N}(i)} \frac{1}{|\mathcal{N}(i)|} \: \mathds{1}[h_j > 0]
\end{equation}
Here, $h_i$ is the representation for the $i^{\text{th}}$ node ($i^{\text{th}}$ row in $H$) and $\mathds{1}[\cdot]$ is the indicator function. The resulting matrix $P$, with the same dimensionality as $H$, houses additional node features that capture information about the feature diversity in every node's neighborhood.
The \textsc{ppv} function does not differentiate neighbours based on relation types as it is applied as a ``post-processing'' step to representations that are the result of relation-specific transformations.

\subsection{Putting it all Together}
Given a KG $\mathcal{G} = (\mathcal{V}, \mathcal{E}, \mathcal{R})$ with entities $v_i \in \mathcal{V}$, edges $(v_i, r, v_j) \in \mathcal{E}$ and relation types $r \in \mathcal{R}$, we construct Euclidean vectors for every entity. These vectors should encapsulate each entity's semantics, as contained in the triples of the KG, such that they can be used as features for machine learning models. 
Note that entities can also be ``literals'' and have associated values. Both \textsc{r-gcn}s and the proposed \textsc{rr-gcn}s, however, regard literals as ordinary nodes.

\begin{algorithm}[h]
\DontPrintSemicolon
\SetKwInput{KwData}{Inputs}
\SetKwInput{KwResult}{Output}
\KwData{relations $\mathcal{R}$, per-relation node neighbours $\mathcal{N}$, embedding size $e$, number of layers $n$, seed $s$, nodes $\mathcal{V}$}
\KwResult{Matrix $\in \mathbb{R}^{|\mathcal{V}| \times e}$}
  \SetKwProg{Fn}{Function}{:}{}
  \SetKwFor{RepTimes}{repeat}{times}{end}
  \Fn{\textsc{embed-rrgcn-ppv}($e$, $s$, $n$)}{
        $H$ $\gets$ \textsc{normal}($|\mathcal{V}| \times e, s, \sigma^{2}=\frac{1}{e}$) \; 
        $S$ $\gets$ \textsc{random-list}($s, \: |\mathcal{R}| + 1$)\; 
        $H$ $\gets$ (\textsc{rrgcn-conv}$(S, e, H))^+$\;
        $P$ $\gets$ \textsc{ppv}($H, \mathcal{N}$)\;
        \RepTimes{n - 1}{
            $H$ $\gets$ (\textsc{rrgcn-conv}$(S, e, H))^+$\;
            $P$ $\gets$ \textsc{rrgcn-conv}($S, e, P$)\;
            $P$ $\gets$ \textsc{ppv}($P, \mathcal{N}$)\;
        }
        \KwRet $[h \:\| \: p]$
  }
  \caption{Generating node embeddings using \textsc{rr-gcn} layers}\label{algo:rrgcn-ppv}
\end{algorithm}
\setlength{\textfloatsep}{5pt}
Algorithm \ref{algo:rrgcn-ppv} illustrates the process of generating these embeddings in pseudocode. For ease of notation we assume that triples are accessible through $\mathcal{N}_r(i)$, which contains the neighbours connected to an entity $v_i$ through a relation $r \in \mathcal{R}$. We also assume that the relations in $\mathcal{R}$ are encoded as integers starting at one. Apart from the KG's characterisation, our method takes an embedding size $e$, a seed $s$, and a number of layers $n$ as input. Based on the seed $s$, random initial embeddings are sampled from a normal distribution with variance chosen such that the sum of a node's embedding is a standard normal random variable. We chose this initialisation strategy over Glorot as the range of Glorot-initialised matrices gets smaller with both the number of rows and columns. This makes sense for transformation matrices, but not for embeddings.
The seed $s$ is then reused to generate a list of additional random seeds, one for every transformation matrix $W_i$. This list of seeds is given as an argument to the \textsc{rrgcn-conv} function, listed in Algorithm \ref{algo:rrgcn-conv}.

\begin{algorithm}[h]
\DontPrintSemicolon
\SetKwInput{KwData}{Inputs}
\SetKwInput{KwResult}{Output}
\KwData{list of seeds $S$, embedding size $e$, node representations $H$}
\KwResult{Matrix $\in \mathbb{R}^{|\mathcal{V}| \times e}$}
  \SetKwProg{Fn}{Function}{:}{}
  \SetKwFor{RepTimes}{repeat}{times}{end}
  \Fn{\textsc{rrgcn-conv}($S$, $e$, $H$)}{
        $X$ $\gets$ $O_{e \times e}$ \;
        \ForEach{$r \in \mathcal{R}$}{
            $W_r$ $\gets$ \textsc{random-uniform}($e \times e,\: S_r$) \;
            $x_i$ $\gets$ $h_i$ + $\sum\limits_{j \in \mathcal{N}_r(i)} \frac{1}{|\mathcal{N}_r(i)|} \, W_r h_j^{(l)}$\;
        }
        $W_0$ $\gets$ \textsc{glorot-uniform}($e \times e,\: S_0$)\;
        $x_i$ $\gets$ $h_i + W_0h^{(l)}_i$\\
        \KwRet $X$
  }
  \caption{\textsc{rr-gcn} Message Passing}
  \label{algo:rrgcn-conv}
\end{algorithm}
\setlength{\textfloatsep}{5pt}

The \textsc{rrgcn-conv} function implements the \textsc{r-gcn}'s message passing equation (see Equation~\ref{eq:methrgcn}) with optimised memory usage. Relation-specific (and self-loop) contributions are iteratively added to the zero-initialised ($O_{e \times e}$) output representation matrix $X$, generating the required transformations on-the-fly from the given seeds. Note that $x_i$ and $h_i$ stand for the $i^{\text{th}}$ row of $X$ and $H$ respectively, and that the computations for these rows are done all at once using efficient matrix multiplications; we only use the row-wise notation for clarity.

After the first message passing round, the obtained representations are passed to a ReLU non-linearity ($^+$) and stored in $H$. These hidden representations are then used to calculate the one-hop \textsc{ppv} features $P$ as described in Section~\ref{subsec:ppv}. The ReLU activation function is not used for \textsc{ppv} representations as they are positive by definition and the result of a non-linear operation.

Using the updated hidden node features, the previous steps are repeated $n - 1$ times. There are two noteworthy details here: (i) we use the same seeds for every convolution, and (ii) we convolute the \textsc{ppv} features independently from the regular representations (but with the same seeds). In our experiments, we noticed that using different transforms in every layer was not necessary -- and sometimes even hurt performance -- for \textsc{rr-gcn}s. 

After the random message passing layers, the concatenation ($\|$) of $H$ and $P$ along the horizontal axis yields the final node embeddings.

\subsection{Memory Usage}
Trained \textsc{r-gcn}s need to store their parameters in (GPU) memory. Since relation-specific transformations are typically quite small, this is dominated by the initial node representations. Especially when using one-hot encodings as initial representations, this memory load can become quite significant.
Additionally, in the backward pass, gradients for all weights (and for optimisers such as Adam~\cite{kingma2014method}, even gradient moments) have to be stored as well, which at least doubles the parameter memory requirements.
For a KG with nodes $\mathcal{V}$, an \textsc{r-gcn} with $B$ bases and an initial representation size $e$ thus needs to be able to keep \textit{at least} two $B \times |\mathcal{V}| \times e$-dimensional float-tensors (4 bytes per element) in memory. For DBLP (a dataset with 4,470,778 nodes, Table \ref{table:statistics2}) and a 40-base \textsc{r-gcn} with a 16-dimensional embedding size this results in at least 22.89GB of memory.

An even bigger source of memory usage for trained \textsc{r-gcn}s, is the storage of intermediate activations. Even if per-relation contributions are iteratively summed to an accumulator as in Algorithm \ref{algo:rrgcn-conv}, by default, all individual contributions accross \textit{all layers} are still kept in memory during training, as they are required for backpropagation. This results in a $|\mathcal{V}| \times e_l$-dimensional float-tensor for \textit{every relation} $r \in \mathcal{R}$ and the layer's global output, for every layer $l$, with $e_l$ the output dimensionality for layer $l$. For DBLP (136 relations with inverses) and a single \textsc{r-gcn} layer with a 16-dimensional output, e.g., this results in 39.20GB. Activation checkpointing~\cite{chen2016training} could be used to avoid storing intermediate activations by recalculating them during the backward pass, trading in memory for compute, but no current implementations use this approach.

Since \textsc{rr-gcn}s do not need a backward pass, we do not need to keep any gradients or intermediate activations in memory. Moreover, because the initial node embeddings and transformation matrices are random, we also do not keep them in memory; storing the seeds is sufficient to recreate the necessary tensors when they are needed. The peak memory usage for an \textsc{rr-gcn} with embedding size $e$ is dominated by (1) the previous-layer node representations, (2) the previous-layer \textsc{ppv} features, (3) the accumulator matrix $X$ (Algorithm \ref{algo:rrgcn-conv}) and (4) the intermediate results for \textit{a single} relation type; four $|\mathcal{V}| \times e$-dimensional float-matrices, which is comparable to the memory complexity of an \textsc{r-gcn}'s forward pass during inference. \textsc{rr-gcn}s clearly need much less memory than their trained counterparts \textit{for a given embedding size}.
However, as these embeddings result from combinations of randomly transformed random representations, to capture useful information, a larger embedding size is needed compared to \textsc{r-gcn}s, which are trained to select \textit{only} the useful features for the downstream task.
As such, \textsc{rr-gcn}s are not necessarily more memory efficient than \textsc{r-gcn}s, depending on how many random features are necessary for the downstream task. For DBLP and an \textsc{r-gcn} with embedding size 512, e.g., the resulting peak memory usage is 36.62GB.
It is important to note that, once trained, \textsc{r-gcn} inference actually requires less memory because of the more compact representations.

\section{Results} \label{sec:results}
In this section, we present empirical results that show how our \textsc{rr-gcn} method matches -- and in some cases exceeds -- the performance of end-to-end trained networks for the two main KG machine learning tasks: node classification and link prediction, which we first briefly introduce along with the datasets used for evaluation. We conclude the section with a visual exploration of the resulting embedding spaces.

\subsection{Node Classification}
For node classification, we are provided with two subsets of vertices $\mathcal{V}_{tr} \subseteq \mathcal{V}$ and $\mathcal{V}_{te} \subseteq \mathcal{V}$ with corresponding labels $y_{tr}$ and $y_{te}$. The goal of this task is to construct a model or hypothesis $h(.)$ based on $\mathcal{V}_{tr}$, $y_{tr}$, and the graph's edges $\mathcal{E}$ that minimises a loss function $\mathcal{L}$(.) and generalises well to the unseen labeled vertices in $\mathcal{V}_{te}$:
\begin{equation}
    \argmin_{h} \mathcal{L}(y_{te}, h_{\mathcal{V}_{tr}, y_{tr}}(\mathcal{V}_{te}))
\end{equation}

\setlength{\tabcolsep}{8pt}
\begin{table}[h!]
\begin{center}
 
\begin{tabular}{lllll}\toprule
 \textbf{Statistic}           & \textbf{AIFB} & \textbf{MUTAG}   & \textbf{BGS} & \textbf{AM}\\\midrule
Entities  &  8,285  & 23,644  & 333,845 & 1,666,764 \\
Relations  & 45  & 23  & 103 & 133  \\
Edges  &  29,043 & 74,227 & 916,199 & 5,988,321 \\
Train Entities  & 141  &  272 & 117 & 802  \\
Val. Entities  & 0 & 0 & 0 & 0  \\
Test Entities  & 37  & 68 & 29 &  198 \\
Classes  & 4  & 2  & 2 & 11 \\
Mean Degree & 7.82 & 6.27 & 11.89 & 13.74 \\
Max. Degree & 1,281 & 6,783 & 83,024 & 73,447 \\ \bottomrule
\end{tabular}
\end{center}
\caption{Dataset statistics for the small-scale KGs from Ristoski et al.\label{table:statistics}}
\end{table}

A first collection of datasets we use to compare \textsc{rr-gcn}s and \textsc{r-gcn}s are those published by Ristoski et al.~\cite{ristoski2016collection}. These four datasets have varying sizes, but most of them are of a rather small scale. We include them because they were used to evaluate node classification in the original \textsc{r-gcn} paper~\cite{schlichtkrull2018modeling}. Table~\ref{table:statistics} summarises some important properties of the four datasets.

\setlength{\tabcolsep}{3pt}
\begin{table}[h!]
\begin{center}
 
\begin{tabular}{llllll}\toprule
 \textbf{Statistic}          &  \textbf{AMPLUS} &  \textbf{DMG777K}   &  \textbf{DMGFULL} &  \textbf{MDGENRE} &  \textbf{DBLP}\\\midrule
Entities  &  1,153,679  & 341,270  & 842,550 & 349,344 & 4,470,778 \\
Relations  & 33  & 60  & 62 & 154 & 68  \\
Edges  & 2,521,046 & 777,124 & 1,850,451 & 1,252,247 & 21,985,048 \\
Train Entities  & 13,423 & 5,394 & 23,566 & 3,846 & 26,535  \\
Val. Entities  & 20,000 & 1,001 & 10,001 & 1,006 & 10,000 \\
Test Entities  & 20,000 & 2,001 & 20,001 & 3,005 & 20,000 \\
Classes  & 8  & 5  & 14 & 12 & 2 \\
Mean Degree & 4.37 & 4.53 & 4.47 & 7.17 & 9.83 \\
Max. Degree & 154,828 & 65,576 & 121,217 & 57,363 & 3,364,084 \\ \bottomrule
\end{tabular}
\end{center}
\caption{Dataset statistics for the larger-scale ``kgbench'' KGs\label{table:statistics2}}
\end{table}

In addition, we compare both techniques on ``kgbench''~\cite{bloem2021kgbench}, which contains five larger scale KGs. Table~\ref{table:statistics2} summarises some important properties of the five datasets. As opposed to the smaller-scale benchmark KGs, these contain a large number of training and testing entities. Moreover, separate validation sets are provided for hyper-parameter tuning.

\subsection{Link Prediction}
The goal of link prediction is to infer true triples that are not yet captured in a KG. To that end, the KG's known triples are split in two subsets $\mathcal{E}_{tr} \subseteq \mathcal{E}$ and $\mathcal{E}_{te} \subseteq \mathcal{E}$ and a model or hypothesis $h(.)$ is constructed. This hypothesis, or ``scoring function'' is trained to assign a high output to the true edges $\mathcal{E}_{tr} \subseteq \mathcal{E}$ and a low output for corrupted, false edges $\mathcal{E}_{tr,c} \not\subset \mathcal{E}$.
The trained hypothesis should then be able to recover the withheld edges $\mathcal{E}_{te}$ by assigning them high scores, and low scores to \textit{all} false edges, i.e., it should minimise a loss function $\mathcal{L}$(.):
\begin{equation}
    \argmin_{h} \mathcal{L}(h_{\mathcal{E}_{tr}}(\mathcal{E}_{te}), h_{\mathcal{E}_{tr}}(\overline{\mathcal{E}_{tr} \cup \mathcal{E}_{te}}))
\end{equation}

\begin{table}[h!]
\begin{center}
 
\begin{tabular}{lllll}\toprule
     \textbf{Statistic}      & \textbf{FB15k-237}\\\midrule
Entities  &  14,541 \\
Relations  & 237 \\
Train Triples  & 272,115  \\
Val. Triples  & 17,535  \\
Test Triples  & 20,466 \\
Mean Degree &  37.52 \\
Max. Degree & 7614 \\ \bottomrule
\end{tabular}
\end{center}
\caption{Dataset statistics for FB15k-237\label{table:statistics3}}
\end{table}

We evaluate our \textsc{rr-gcn}'s link prediction performance only on the FB15k-237 dataset, as the other datasets used for evaluation in the original \textsc{r-gcn} work~\cite{schlichtkrull2018modeling} have all since been found to contain significant leakage~\cite{dettmers2018convolutional}. Table~\ref{table:statistics3} summarises some important properties of this dataset.

\subsection{Evaluation} \label{sec:setup}
We implemented our \textsc{rr-gcn} layer and embedder in PyG~\cite{NEURIPS2019_9015}, an extension of the popular deep learning framework PyTorch~\cite{Fey/Lenssen/2019} that facilitates the implementation of message passing networks. PyG provides parallel execution of node representation updates. In the remainder of this section, we discuss the evaluation setup for the node classification and link prediction tasks, and introduce our methodology for a rudimentary qualitatitve analysis of the embeddings.

\subsubsection{Node Classification}
A first step in our evaluation procedure is to reduce the size of the KG by excluding vertices that are further than $n$ hops away from any of the training ($\mathcal{V}_{tr}$) or testing ($\mathcal{V}_{te}$) vertices, as no information from more than $n$ hops away can be propagated to the nodes of interest with $n$ message passing layers. Once the size of the KG is reduced, we apply $n$ layers of our \textsc{rr-gcn} to create embeddings of size $e$. The constants $n$ and $e$ are tuneable hyper-parameters. Once the embeddings are generated, they are provided as input to a gradient boosting classifier. In this study, we used CatBoost~\cite{dorogush2018catboost}.

For the small-scale benchmark KGs, the hyper-parameters $n$ and $e$ were tuned using a grid search with stratified 5-fold cross-validation, with a different \textsc{rr-gcn} seed in every fold. For the larger-scale KGs, we evaluated an \textsc{rr-gcn} five times with different seeds on the provided validation set to tune the hyperparameters. We tuned $n$ to take a value in $\{1, 2, 3, 4, 5\}$ and $e$ to take a value in $\{256, ~384, ~512, ~768, ~1024 \}$ and chose optimal values based on log-loss, ignoring configurations that resulted in more than 24GB of GPU memory (which corresponds to the current most high-end consumer GPU, the RTX 3090). 
An important hyperparameter for CatBoost is the number of boosting iterations. During validation, we determined this quantity using ``early stopping''. For evaluation runs on a given dataset's test set, we chose the maximum number of iterations required for that dataset's validation runs with optimal hyperparameters. 

\begin{table}[t]
\begin{center}
 
\begin{tabular}{lcccc}
\toprule
{} & \multicolumn{2}{c}{\textbf{number of layers}} & \multicolumn{2}{c}{\textbf{embedding size}} \\\cmidrule(lr){4-5} \cmidrule(lr){2-3}
 \textbf{Model} &      \textsc{no ppv} &  \textsc{ppv}  &         \textsc{no ppv} &     \textsc{ppv} \\
\midrule
AIFB    &      4 &  1 &       256 &   512 \\
AM      &      5 &  5 &       768 &   768 \\
BGS     &      5 &  5 &       512 &   512 \\
MUTAG   &      2 &  2 &      1024 &  1024 \\\midrule
AMPLUS  &      5 &  5 &      1024 &  1024 \\
DBLP    &      5 &  5 &       256 &   256 \\
DMG777K &      2 &  2 &      1024 &  1024 \\
DMGFULL &      2 &  1 &       256 &  1024 \\
MDGENRE &      5 &  5 &       768 &  1024 \\
\bottomrule
\end{tabular}
\end{center}
\caption{Optimal hyperparameters per dataset, for \textsc{rr-gcn}s that use \textsc{ppv} features and \textsc{rr-gcn}s that leave them out \label{table:hyperparams}}
\end{table}

We reproduced \textsc{r-gcn} results from Schlichtkrull et al.~\cite{schlichtkrull2018modeling} using an external implementation~\cite{thanapalasingam2021relational}, as the original code uses deprecated libraries. For the small-scale KGs, we used the hyper-parameter configuration reported in their study~\cite{schlichtkrull2018modeling}, but we additionally used early stopping with $10$ epochs patience using a validation set held out from the training data to determine the optimal number of epochs. 
For the larger-scale KGs, we use the hyper-parameters as reported in the ``kgbench'' paper~\cite{bloem2021kgbench}. All ``kgbench'' measurements were performed on CPU, to accommodate the high memory requirements for backpropagation. Only for the DBLP dataset, for which no results were reported in the original paper due to high memory requirements, we had to use an embedding size of $10$ (as opposed to $16$) with $40$ base functions to make it fit in our 64GB of CPU RAM. We used an early stopping mechanism, with $10$ epochs patience, using the provided validation sets.

We evaluated the \textsc{r-gcn} and \textsc{rr-gcn} setups ten times with different seeds and measured the according test accuracies. We report results for both the \textsc{rr-gcn} as described in Algorithm~\ref{algo:rrgcn-ppv} (\textsc{rr-gcn-ppv}) and a version that does not include the \textsc{ppv} features (\textsc{rr-gcn}). The mean accuracy results, and their corresponding standard errors are provided in Table~\ref{table:accuracies}.

\begin{table}[t]
\begin{center}
\begin{tabular}{llll}\toprule
   \textbf{Dataset}   & \textsc{\textbf{r-gcn}}                 & \textsc{\textbf{rr-gcn}}               & \textsc{\textbf{rr-gcn-ppv}} \\\midrule
AIFB & \bftab 96.11 $\pm$ 0.45 & 83.33 $\pm$ 1.37 & 86.11 $\pm$ 0.93 \\
AM & \bftab 88.99 $\pm$ 0.39 & 81.67 $\pm$ 0.57 & 84.65 $\pm$ 0.62 \\
BGS & \bftab 86.21 $\pm$ 0.89 & 80.00 $\pm$ 2.34 & 78.97 $\pm$ 2.44 \\
MUTAG & 72.50 $\pm$ 0.91 & 70.00 $\pm$ 0.83 & \bftab 79.41 $\pm$ 0.58 \\
\midrule
AMPLUS & 83.81 $\pm$ 0.13 & 76.85 $\pm$ 0.06 & \bftab 84.54 $\pm$ 0.08 \\
DBLP & 68.51 $\pm$ 0.99 & 70.18 $\pm$ 0.11 & \bftab 70.61 $\pm$ 0.07 \\
DMG777K & 62.51 $\pm$ 0.38 & 61.40 $\pm$ 0.32 & \bftab 63.97 $\pm$ 0.26 \\
DMGFULL & 57.52 $\pm$ 0.19 & 60.50 $\pm$ 0.26 & \bftab 63.38 $\pm$ 0.17 \\
MDGENRE & \bftab 67.33 $\pm$ 0.19 & 65.09 $\pm$ 0.10 & 67.15 $\pm$ 0.08 \\\bottomrule
\end{tabular}
\end{center}
\caption{The average accuracy and standard error of our $10$ measurements.\label{table:accuracies}}
\end{table}

\subsubsection{Link Prediction}
As in the original work on \textsc{r-gcn}s~\cite{schlichtkrull2018modeling}, we generated false triples using ``negative sampling'', where the head or tail of each true triple in $\mathcal{E}_{tr}$ is randomly corrupted, i.e., replaced by another entitiy. We trained a scoring function (or ``decoder'') using the binary cross-entropy loss to score true triples higher than negatives. Whereas \textsc{r-gcn}s use the bilinear DistMult decoder~\cite{schlichtkrull2018modeling,yang2015embedding} for link prediction, we preceded DistMult by a small three-layer neural network with hidden and output size 2048, to allow the decoder to transform our task-agnostic embeddings into a space in which the bilinearity holds. 

As trained \textsc{r-gcn}s require much larger hidden sizes for link prediction than for node classification (500 instead of 16 in the original work), our random unsupervised embeddings need to be very large to capture a similar amount of information. As such, we used an \textsc{rr-gcn-ppv} with embedding size 32,000. Using principal component analysis (after feature normalisation), we reduced the representation dimensionality to 8,192 before feeding them to a three-layer pre-processing neural network that was trained jointly with the 2048-dimensional DistMult decoder. We determined the number of epochs to train the decoder using the provided validation set.

Since we used the same final decoding assumption, and training \textsc{r-gcn}s for link prediction can take \textit{several days}~\cite{thanapalasingam2021relational}, for the end-to-end model, we reused the results reported in~\cite{schlichtkrull2018modeling}. The filtered mean reciprocal rank (FMRR) and the hits at 1, 3 and 10 are listed Table~\ref{tab:linkpred}. 

\begin{table}[t]
\begin{center}
\begin{tabular}{llllll}\toprule
 \textbf{Model}      &  \textbf{FMRR} &  \textbf{H@1} &  \textbf{H@3} & \textbf{H@10}\\\midrule
\textsc{r-gcn} &  0.248 & 0.153 & 0.258 & 0.414\\
\textsc{rr-gcn-ppv} & 0.238 & 0.157 & 0.256 & 0.412  \\\bottomrule
\end{tabular}
\end{center}
\caption{Link prediction results for FB15k-237 of \textsc{r-gcn} and our proposed \textsc{rr-gcn}.\label{tab:linkpred}}
\end{table}

\subsubsection{Qualitative Evaluation} \label{sec:qual_evaluation}
\begin{figure}[h!]
    \centering
    \includegraphics[width=0.9\textwidth]{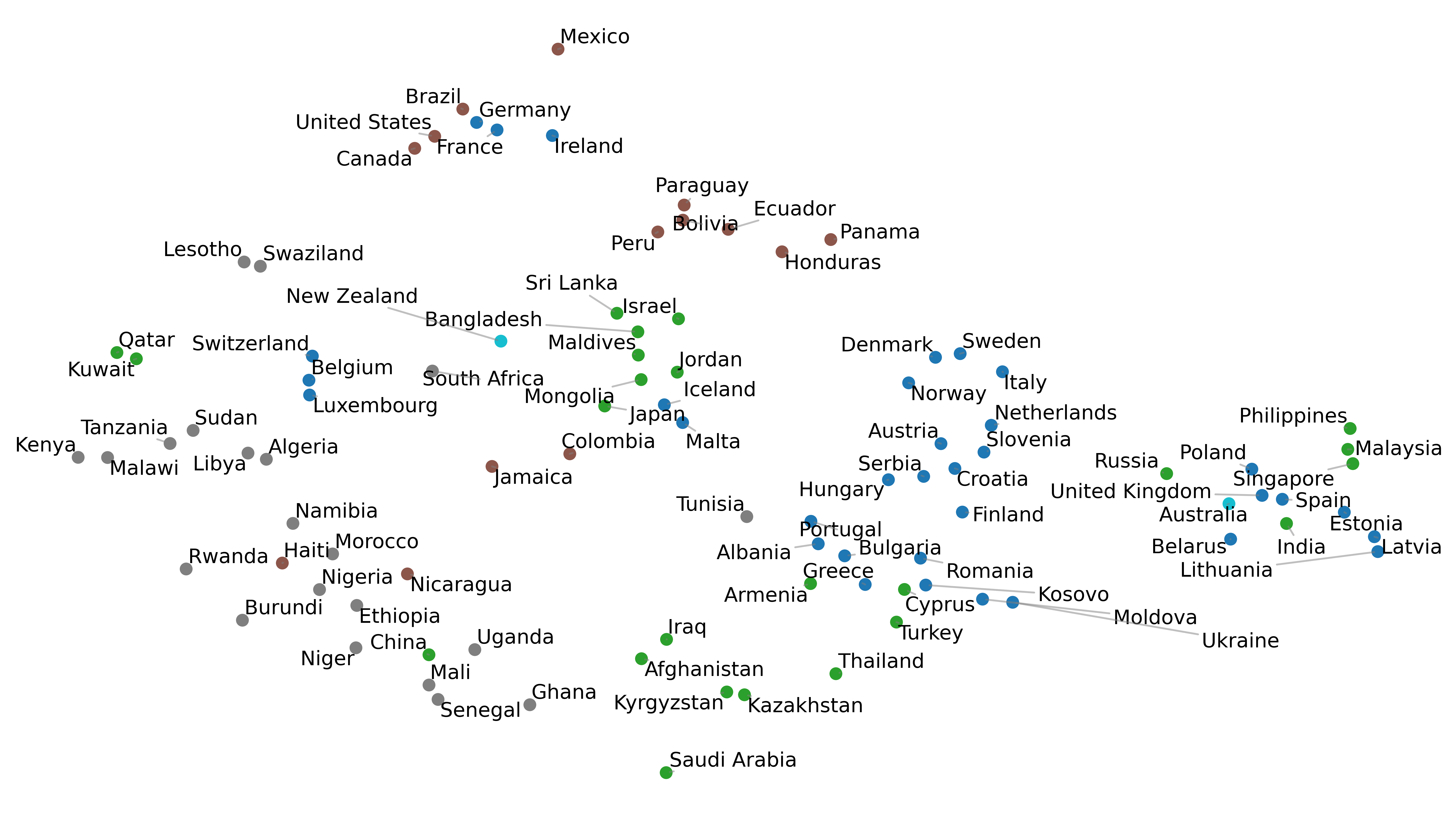}
    \caption{t-SNE representation of the embeddings for countries in a subset of DBpedia produced by \textsc{rr-gcn}.}
    \label{fig:t-sne-countries}
\end{figure}

We perform a qualitative evaluation of our produced \textsc{rr-gcn} embeddings by visualising \textsc{rr-gcn} representations for countries in a small subset of DBpedia~\cite{dbpedia,dbpedia_countries} to gauge if the resulting embedding space makes semantic sense. Figure \ref{fig:t-sne-countries} shows a plot of t-SNE-transformed~\cite{van2008visualizing} embeddings generated by a 5-layer \textsc{rr-gcn-ppv} with embedding size 512.

\section{Discussion} 
Table~\ref{table:accuracies} shows that, for most small-scale datasets, \textsc{r-gcn}s score better. However, the differences are not that drastic considering one method is trained end-to-end and another is unsupervised and based on \textit{random} transformations.
\textsc{ppv} features, which measure the ``representation diversity'' around every node, seem to have a positive impact for most datasets. With \textsc{ppv}, our random \textsc{rr-gcn}s even score significantly better for the MUTAG dataset. Only for BGS, \textsc{ppv} seemingly detoriates performance. However, the difference between \textsc{rr-gcn-ppv} and \textsc{rr-gcn} is statistically insignificant (Mann-Whitney test with cutoff $0.05$).

For the larger datasets, the average performance of \textsc{r-gcn}s and \textsc{rr-gcn}s is even more similar, with most datasets statistically significantly favouring \textsc{rr-gcn-ppv} over a trained model. The difference for MDGENRE, the only ``kgbench'' dataset where \textsc{r-gcn}s score better on average, is statistically insignificant. 

The KGs on which \textsc{rr-gcn} performs significantly worse than \textsc{r-gcn} often contain many low-degree nodes (including literals, which usually have only one neighbour) close to the nodes to be classified. We hypothesise that these low-degree nodes add noise to our representations, which can overpower the useful signal in some datasets.
As \textsc{r-gcn}s can learn initial node embeddings and per relation transformations, they can learn to ignore unnecessary information, or at least make the useful signal more influential in the final representations. To test this hypothesis, we apply \textsc{r-gcn} and \textsc{rr-gcn-ppv} to filtered versions of AIFB, BGS and AM. In these filtered KGs, low-degree nodes (degree $\leq$ 5) are removed. 
\begin{table}[t]
\begin{center}
\begin{tabular}{lllll}\toprule
\textbf{Model}      &  \textbf{AIFB} &  \textbf{BGS} &  \textbf{AM} \\\midrule
\textsc{r-gcn} & \bftab 96.11 $\pm$ 0.45 & \bftab 86.21 $\pm$ 0.89 & \bftab 88.99 $\pm$ 0.39 \\
\textsc{rr-gcn-ppv} & 86.11 $\pm$ 0.93 & 78.97 $\pm$ 2.44 & 84.65 $\pm$ 0.62 \\\midrule
\textsc{r-gcn-cut} & 95.56 $\pm$ 0.45 & \bftab 86.21 $\pm$ 0.89 & \bftab 88.13 $\pm$ 0.41 \\
\textsc{rr-gcn-ppv-cut} & \bftab 95.83 $\pm$ 0.62 & 84.14 $\pm$ 1.38 & 84.80 $\pm$ 0.23 \\\bottomrule

\end{tabular}
\end{center}
\caption{The average accuracies and standard error of our $10$ measurements with degree cutting.\label{table:degreecutaccuracies}}
\end{table}

As shown in Table~\ref{table:degreecutaccuracies}, removing low-degree nodes seems to close the gap in average performance for most datasets: the performance for \textsc{r-gcn} stays roughly the same, while \textsc{rr-gcn} results improve. The performance differences to the trained counterparts for AIFB and BGS even become statistically insignificant. For AM, however, low-degree node removal seems to have little to no effect. As we suspect that, for this dataset, the \textsc{r-gcn} learns to ignore a subset of relation types, we inspect the mean absolute values of a trained \textsc{r-gcn}'s \textit{learned} per-relation transformation matrices (true and added inverse relation types are averaged), averaged over all layers. 
\begin{figure}[t]
    \centering
    \includegraphics[width=0.6\textwidth]{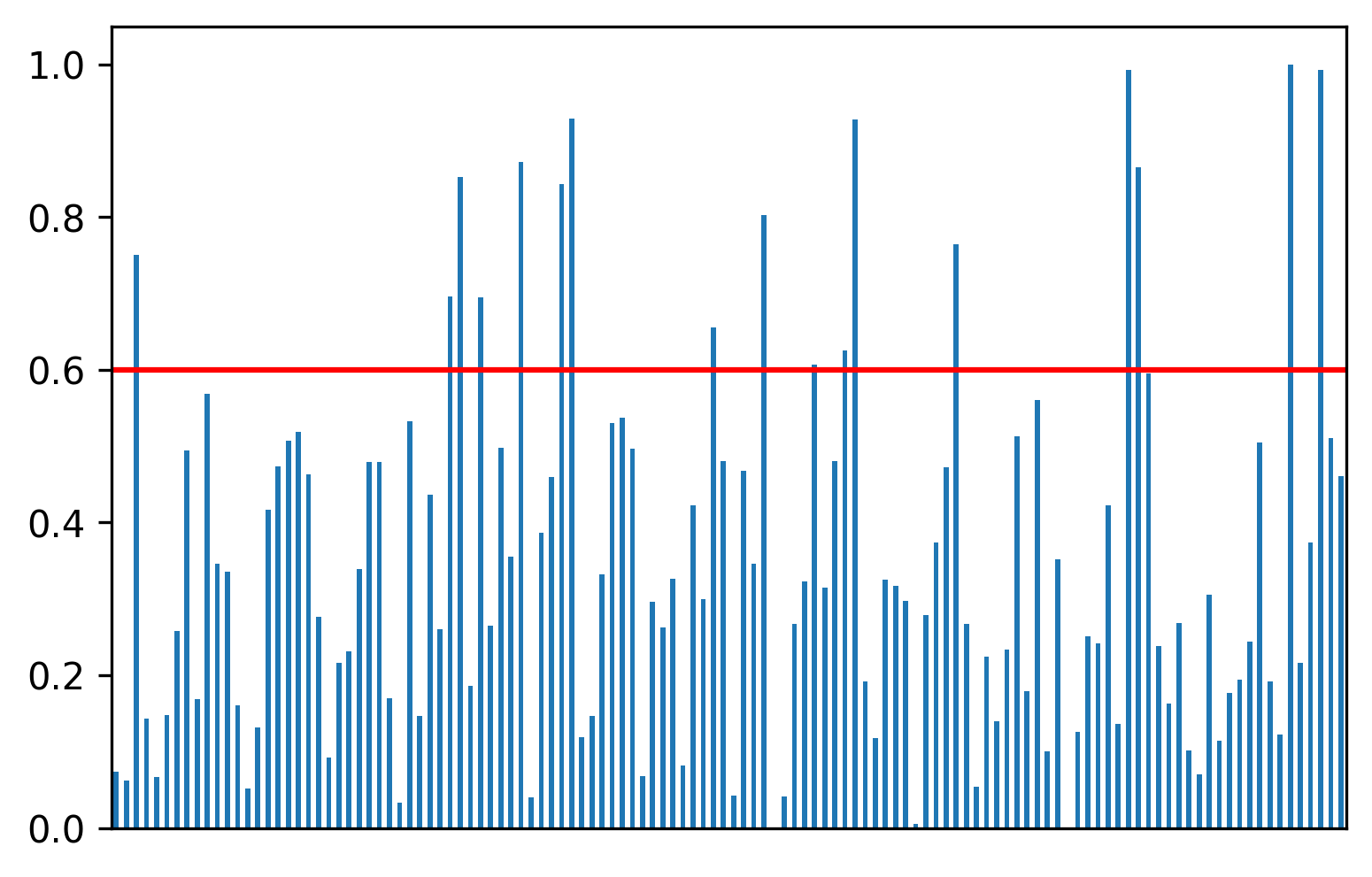}
    \caption{Mean absolute weights for per-relation transformations learned by a two-layer \textsc{r-gcn} for the AM dataset, relative to the maximum per-relation mean absolute weight.}
    \label{fig:amimportances}
\end{figure}
Indeed, Figure \ref{fig:amimportances} illustrates that there are quite a few relation types that a trained \textsc{r-gcn} learns to attenuate. If we take this ``relation importance'' information into account by removing (in addition to low-degree nodes) all relation types that have less than 60\% of the maximum ``importance'', our \textsc{rr-gcn} obtains an accuracy of 91.31 $\pm$ 0.24\%, which is even statistically significantly better than trained \textsc{r-gcn}s.

Table \ref{tab:linkpred} illustrates that link prediction performance for random and trained \textsc{r-gcn}s is similar across all evaluation metrics. Which indicates that, even when \textsc{r-gcn}s need large hidden sizes to perform downstream tasks (500 in this case), random message passing can capture a comparable amount of signal, without supervision. As only the decoder needs to be trained, the \textsc{rr-gcn} encoder needs to run only once. This results in a significant speedup compared to end-to-end trained \textsc{r-gcn}s, which have to perform both a forward and a backward pass trough the entire network for every training step. Consequently, the \textsc{rr-gcn} link prediction model trains in just under two hours, while \textsc{r-gcn}s for this task can take days to converge~\cite{thanapalasingam2021relational}.

Figure \ref{fig:t-sne-countries} shows that \textsc{rr-gcn} embeddings preserve semantic relationships between entities: there is a quite clear separation between continents, neighbouring countries are generally close to each other, and some countries with intertwined geopolitical histories even form small clusters.

\section{Conclusion} \label{sec:conclusion}
Inspired by the success of random non-linear transformations in the time series domain, we set out to evaluate random convolutions (\textsc{rr-gcn}) on KGs. In this exploratory study, empirical results show that random transformations usually match or exceed the performance of end-to-end trained \textsc{r-gcn}s. 
Our experiments indicate that a KG's \textit{structure} alone, which is the only thing our \textsc{rr-gcn}s can capture, contains enough semantic information as is, and that when \textsc{r-gcn}s do perform better, they mostly learn to \textit{ignore} certain parts of that structure.

Aside from surprisingly good performance, \textsc{rr-gcn}s exhibit other interesting properties: (i) embeddings can easily be generated for a small subset of the graph for testing purposes (whereas \textsc{r-gcn}s need to process the entire graph to train their weights), and (ii), as training through backpropagation is not necessary, they can be used to test the influence of new e.g.~aggregation functions (such as \textsc{ppv}) before considering differentiable variants. This opens up many new interesting avenues for further research, both in improving the \textsc{rr-gcn}s themselves and KG embedding in general, which we discuss subsequently.

\section{Future Work}
\label{sec:futurework}
Apart from being used as a baseline that any trained method should be able to beat, we envision that \textsc{rr-gcn}s might be helpful as a tool for improving and/or developing other embedding techniques. By further investigating in which situations or for which substructures random information aggregations are insufficient, trained algorithms could be altered to better target problematic computations. Hybrid networks with both trained and random weights could offer the best of both worlds.

In this work we explored information aggregation using random transformation matrices. Other strategies, e.g.~the integration of algebraic embedding priors into message passing~\cite{2019transgcn,compgcn}, could be randomised as well.

As it gives a measure of the representation diversity around a node and thus captures more of the KG's structure, \textsc{ppv} seems like promising addition to the random message passing toolbox. However, the independent computation of \textsc{ppv} and regular features is somewhat arbitrary. Future work could research other options such as calculating \textsc{ppv} features only after the last message passing layer, or even outright replacing the averaging used in \textsc{r-gcn}s and using \textsc{ppv} as the aggregation. Moreover, differentiable alternatives to \textsc{ppv} could be researched to improve trained \textsc{r-gcn}s.
Future research might reveal new features that capture other aspects of graph, to further improve performance.

Even though for some datasets (e.g.~DBLP), \textsc{rr-gcn}s require less memory than their trained counterparts for comparable performance, the memory requirements are still substantial for graphs with many nodes. Future research could trade in some of these memory requirements for compute by e.g.~aggregating many small node embeddings.

\paragraph*{Supplemental Material Statement:} Source code for the \textsc{rr-gcn} embedder and all experiments is attached with the submission on EasyChair and, if accepted, will be published on GitHub.

%
\newpage
\bibliographystyle{unsrt}
\bibliography{main}

\begin{thebibliography}{10}
\bibitem{kipfblog}
How powerful are graph convolutional networks?
\end{thebibliography}

\end{document}